\documentclass{article}

    \PassOptionsToPackage{numbers, compress}{natbib}

\usepackage[preprint]{neurips_2020}

\usepackage[utf8]{inputenc} 
\usepackage[T1]{fontenc}    
\usepackage{hyperref}       
\usepackage{url}            
\usepackage{booktabs}       
\usepackage{amsfonts}       
\usepackage{nicefrac}       
\usepackage{microtype}      
\usepackage{setspace, longtable, graphicx, hyphenat, hyperref, fancyhdr, ifthen, enumitem, amsmath, setspace}
\usepackage{float}
\usepackage{algorithm}
\usepackage{algpseudocode}
\usepackage{multirow}

\title{3FM: Multi-modal Meta-learning for Federated Tasks}

\author{
  Minh Tran\thanks{These authors contributed equally to this work.}\\
  Robotics Institute \\
  Carnegie Mellon University\\
  \texttt{minht@cs.cmu.edu} \\
  \And
  Roochi Shah\footnotemark[1] \\
  Department of Statistics and Data Science \\
  Carnegie Mellon University \\
  \texttt{rtshah@andrew.cmu.edu} \\
  \And
  Zejun Gong\footnotemark[1] \\
  Department of Electrical and Computer Engineering \\
  Carnegie Mellon University\\
  \texttt{zejung@andrew.cmu.edu} \\
  }

\begin{document}
\maketitle
\begin{abstract}
We present a novel approach in the domain of federated learning (FL), particularly focusing on addressing the challenges posed by modality heterogeneity, variability in modality availability across clients, and the prevalent issue of missing data. We introduce a meta-learning framework specifically designed for multimodal federated tasks. Our approach is motivated by the need to enable federated models to robustly adapt when exposed to new modalities, a common scenario in FL where clients often differ in the number of available modalities. The effectiveness of our proposed framework is demonstrated through extensive experimentation on an augmented MNIST dataset, enriched with audio and sign language data. We demonstrate that the proposed algorithm achieves better performance than the baseline on a subset of missing modality scenarios with careful tuning of the meta-learning rates. This is a shortened report, and our work will be extended and updated soon. Code and full report could be found on our \href{https://github.com/minhtcai/MLMF}{Github}. 
\end{abstract}

\section{Introduction}
Federated learning, \cite{fedavg}, a paradigm of distributed machine learning, faces unique challenges when extended to multimodal data. The heterogeneity of modalities, coupled with their variable availability across different clients, and the frequent occurrence of missing data, poses significant obstacles. Previous work in multimodal learning assumes full availability of all modalities. When applied to the federated setting, this assumption falters as systems heterogeneity challenges cannot guarantee full availability of all modalities across every client. This paper introduces a meta-learning based approach tailored to multi-modal federated tasks. This approach is designed to collaboratively learn a global model in spite of missing data in one or multiple modalities across clients. Our work is inspired by previous works applying meta-learning to overcome missing data in various modalities and applying meta-learning to overcome systems heterogeneity challenges in federated learning.

\section{Related Works}

\subsection{Federated Meta Learning}
Federated learning, known for its decentralized nature and emphasis on privacy, faces challenges like data heterogeneity and communication overhead. Meta-learning, or learning to learn, offers solutions by enabling models to quickly adapt to new data distributions. FedMeta, \cite{fedmeta}: As previously discussed, FedMeta demonstrates how integrating meta-learning in FL can lead to enhanced communication efficiency, faster convergence, and improved accuracy while maintaining privacy. Meta-learning approaches such as Model Agnostic Meta-Learning \cite{maml} have shown promise to improve personalization when applied to federated settings \cite{personalized_fed} by allowing models to adapt rapidly with minimal data. This synergy of FL and meta-learning is key in developing robust and efficient learning systems.

\subsection{Federated Multimodal}
Federated learning in multimodal contexts addresses challenges of learning from diverse data types (like images, text, audio) across distributed nodes. This field is complex due to the varying nature and availability of modalities at different nodes, which is a challenge we seek to address in our work. FedMultimodal, \cite{multifed}: FedMultimodal is a framework which assesses FL robustness against multimodal data issues and proposes a systematic approach for multimodal FL. Other contributions in this domain often focuses on efficient data representation and fusion techniques. For example, MultiModal Federated Learning \cite{multifed} (MMFL)propose methods for integrating diverse data types efficiently in a federated manner.

\subsection{Multimodal Meta Learning}
Meta-learning applied to multimodal learning, \cite{metamul} focuses on creating models that can swiftly adapt to new modalities or changes in data distribution. This is crucial in domains where data from different sources or modalities must be integrated seamlessly. SMIL, \cite{smil}: This paper is a significant contribution to the field, addressing the challenge of missing modalities in multimodal datasets through Bayesian Meta-Learning.



\section{Proposed Approach}
\label{headings}
Our proposed method applies meta-learning principles within a federated learning framework to address the challenges posed by multimodal data diversity within each client. The core of this approach lies in creating a robust initial model that can adapt effectively with the introduction of each new modality. 

As shown in \textbf{Algorithm 1}: for each client, the meta-learning setting is MAML, \cite{maml}, where experiments were design for the model to learn with a limited set of modalities. This aligns with our assumption that not all clients have same number of available modalities. For client training, the available data were splitted to support set and query set as input for the inner loop and outer loop of MAML. The parameters of global model then was updated using FedAvg, \cite{fedavg} for each communication round.

\begin{algorithm}
\caption{FedMeta-Multi-Modal}
\begin{algorithmic}[1]
\State \textbf{// Run on the server}
\State \textbf{AlgorithmUpdate:}
\State Initialize model parameter $\theta$ for MAML.
\For{each communication round $t = 1, 2, \dots$}
    \State Sample a set $U_t$ of $m$ clients, and distribute $\theta$ to the sampled clients
        \For{each client $u \in U_t$ in parallel}
        \State Get test loss $g_u \leftarrow \text{LocalTraining}(\theta)$       
        \EndFor
\EndFor
\State Update algorithm parameters $\theta \leftarrow \theta - \beta \sum_{u \in U_t} g_u$
\State
\State \textbf{// Run on client u}
\State \textbf{LocalTraining}($\theta$):
\State Sample data points \{$x^1_S, x^2_S, x^3_S, y_S$\} $\sim D^u_S$ as support set $D^u_S$ and 
\State \{$x^1_Q, x^2_Q, x^3_Q, y_Q$\} $\sim D^u_S$ as query set $D^u_Q$, the modalities in each sample data point are aligned
\State $L_{D^u_S}(\theta) \leftarrow \frac{1}{|D^u_S|} \sum_{(x,y) \in D^u_S} \ell(f_\theta(x), y)$
\State $\theta^u \leftarrow \theta - \alpha \nabla L^{D^u_S}(\theta)$
\State $L_{D^u_Q}(\theta_u) \leftarrow \frac{1}{|D^u_Q|} \sum_{(x_0,y_0) \in D^u_Q} \ell(f_{\theta_u}(x_0), y_0)$
\State $g_u \leftarrow \nabla_\theta L_{D^u_Q}(\theta_u)$
\State Return $g_u$ to server
\end{algorithmic}
\end{algorithm}

\section{Experiments \& Results}
\subsection{Dataset}
We manually collected 3 independent single-modality dataset and preprocessed them by matching data points with the same label across different datasets to form our multi-modal dataset that contains 2062 samples with 10 labels in total.
\begin{itemize}
    \item \textbf{The Sign-Language-Digits-Dataset} \cite{sign} is a collection of images used for recognizing numerical digits from 0 to 9 in sign language. Each gesture is a distinct representation of a numerical digit. The dataset contains 2062 samples and 10 labels.
    \item \textbf{The MNIST-Dataset} \cite{mnist} consists of 60,000 handwritten training digit images. Each image has a label that represents a digit from 0 to 9.
    \item \textbf{The Free-Spoken-Digit-Dataset} \cite{fsdd} is a simple audio dataset similar in spirit to the classic MNIST dataset but for speech recognition. It contains recordings of spoken digits in English, from 0 to 9. Each digit is pronounced by a variety of speakers. The dataset audios are represented as spectrograms contains 3000 samples in total.
\end{itemize}
\subsection{Network Architecture}
We designed a Multi-Modal Neural Network for processing data from three distinct modalities: image, spectrogram, and sign language. It employs a modular branch structure for each modality, followed by a unified classification stage
We compare the proposed approach with the following baseline method:
Image Branch:
\begin{itemize}
    \item \textbf{The Image branch} Comprises two convolutional layers, each followed by ReLU activation and max-pooling.
    \item \textbf{The Spectrogram Branch} Contains four convolutional layers, each accompanied by ReLU activation. The last two layers are followed by max-pooling.
    \item \textbf{The Sign Branch} structures similarly to the Image Branch with two convolutional layers, ReLU activations, and max-pooling.
\end{itemize}
After processing through their respective branches, the outputs are flattened and concatenated. This combined feature vector is then fed into two fully connected layers for final classification, accommodating up to 10 classes.
\subsection{Baseline Method}
We conduct our baseline experiments by simply training the dataset with missing modalities on the network structure and test their performance on full modality scenarios. For instance, we deliberately mute the sign branch (set its branch output to be zeros) and feed the training data to the network to mimic the scenario of missing the sign modality. Then we test the model performance by bringing back the sign branch and see how the results go for full modality. In Table 1 we show the testing results of performance of training on 6 different missing modality scenarios (column) and testing against full modality (row). 
\begin{table}[ht]
\centering
\begin{tabular}{|l|l|l|l|l|l|l|}
\hline
 & \textbf{img/sign} & \textbf{sp/sign} & \textbf{img/sp} & \textbf{img} & \textbf{spectrogram} & \textbf{sign} \\ \hline
\textbf{img/spectrogram/sign} & 100\% & 20.603\% & 100\% & 100\% & 8.495\% & 13.349\% \\ \hline
\end{tabular}
\caption{Results for training with missing modalities and test on full modality}
\label{tab:your_table_label}
\end{table}
\subsection{3MF method}
We conduct experiments for the proposed 3MF method by configuring 5 local training epochs and 50 global epochs in total. Within each client we split the client dataset into 20\% of support set and 80\% of query set and tested various combinations of outer learning rate and inner learning rate. Similar to the baseline method, we conduct 6 experiments with distinct missing modality scenarios. In each experiment, the support set of each client contains the missing modality data while the query set contains full modality. Note that for each experiment the support set for each client contains the same modality type, each client only has image and spectrogram modality for example. In all experiment settings we set the inner learning rate to be smaller than outer learning rate because the support set size is generally smaller than the query set size. A larger inner learning rate might result in overshooting the optimal parameters for the specific task, leading to poor performance on that task. It might also lead to over-fitting the task for missing the nuanced patterns that are essential for generalization. In table 2 we show the testing performance of the 6 experiment settings.

We observe that our proposed algorithm has the best testing performance on full modality data when client number is set to 3 and outer lr set to 0.001, inner lr set to 0.00001. The model has significant improvement when the clients only contain spectrogram/sign, spectrogram, and sign modality data. Even the rest of the missing modality scenarios does not reach 100\% accuracy as the baseline, we observe smaller variance in performance for all scenarios. For client number= 5 overall, we observe worse performance than the baseline.
We also included the training loss/accuracy curve for the best case in the appendix.

\begin{table}[H]
\centering
\begin{tabular}{|l|l|l|l|l|l|l|l|}
\hline
\multicolumn{2}{|c|}{\textbf{Meta Learning rate}} & \multicolumn{1}{c|}{\multirow{2}{*}{\textbf{img/sign}}} & \multicolumn{1}{c|}{\multirow{2}{*}{\textbf{spect/sign}}} & \multicolumn{1}{c|}{\multirow{2}{*}{\textbf{img/spect}}} & \multicolumn{1}{c|}{\multirow{2}{*}{\textbf{img}}} & \multicolumn{1}{c|}{\multirow{2}{*}{\textbf{spect}}} & \multicolumn{1}{c|}{\multirow{2}{*}{\textbf{sign}}}\\
\cline{1-2}
\textbf{outer lr} & \textbf{inner lr} & & & & & & \\ 
\hline
\multicolumn{8}{|c|}{\textbf{Client Number = 3}} \\ \hline
0.001 & 0.00001 & 86.407\% & \textbf{94.660\%} & 91.747\%  & 94.174\%  & \textbf{69.417\%}  & \textbf{92.718\%}\\
0.01 & 0.00001 & 9.223\% & \textbf{49.757\%} & 9.223\%  & 9.223\%  & \textbf{9.223\%}  & \textbf{40.776\%}\\
0.001 & 0.0001  & 50.000\% & 11.407\% & 92.233\%  & 93.203\%  & \textbf{10.194\%}  & \textbf{83.009\%}\\
0.01 & 0.0001  & 71.475\% & \textbf{80.660\%} & 90.445\%  & 94.174\%  & \textbf{50.417\%}  & \textbf{89.783\%}\\ \hline
\multicolumn{8}{|c|}{\textbf{Client Number = 5}} \\ \hline
0.001 & 0.00001 & 92.233\% & \textbf{65.776\%} & 80.825\%  & 10.679\%  & 7.524\%  & \textbf{70.873\%}\\
0.01 & 0.00001 & 8.737\% & 8.737\% & 8.737\%  & 8.737\%  & \textbf{36.893\%}  & 8.737\%\\
0.001 & 0.0001  & 8.009\% & 8.009\% & 84.223\%  & 8.009\%  & 8.009\%  & \textbf{92.475\%}\\
0.01 & 0.0001  & 8.131\% & 13.132\% & 9.634\%  & 9.152\%  & \textbf{9.143\%}  & 10.512\%\\ \hline
\multicolumn{8}{|c|}{\textbf{Client Number = 10}} \\ \hline
0.001 & 0.00001 & 99\% & \textbf{73.359\%} & 36.523\%  & 83.582\%  & 8.394\%  & \textbf{98.436\%}\\
0.01 & 0.00001 & 9.142\% & 12.132\% & 9.723\%  & 11.342\%  & \textbf{10.412\%}  & 10.452\%\\
0.001 & 0.0001  & 93.203\% & \textbf{98.543\%} & 88.592\%  & 90.776\%  & \textbf{96.601\%}  & \textbf{95.873\%}\\
0.01 & 0.0001  & 30.097\% & \textbf{39.805\%} & 38.834\%  & 27.669\%  & \textbf{9.223\%}  & \textbf{40.291\%}\\ \hline
\end{tabular}
\caption{Performance Comparison Across Different Models and Supports}
\label{tab:performance_comparison}
\end{table}

\section{Limitation and Future directions}
This paper contributes to the burgeoning field of multimodal federated learning by introducing a novel meta-learning framework to learn a global model in spite of missing modalities between clients. Our approach addresses the critical challenges of modality heterogeneity and variability in federated environments. Through extensive experimentation on an augmented MNIST dataset, enriched with audio and sign language, we demonstrate the framework's efficacy in enhancing adaptability to varying modalities. 

\subsection*{Limitations and Future Directions}

\textbf{Limitations}: One limitation of our study is the reliance on a specific dataset, which may not fully represent the diverse range of real-world scenarios. This limitation is in-part caused by computational and time constraints imposed on us through the project (i.e. lack of AWS resources caused by limit increase requests). As a result, we were limited by multimodal datasets that can fit in google colab. Another limitation was the inability to test on the SMIL bayesian meta-learning baseline and integrating it into the federated learning framework. This is due to, in-part, to the complexity of integrating this framework to a federated setting due to ambiguities in the problem formulation, but also due to AWS resource constraints on the increased time it takes to run these experiments. 

\textbf{Future Directions}: Future research should focus on extending the framework to more diverse and complex datasets, exploring the scalability of the approach, and analyzing how its privacy guarantees compare to other federated multimodal baselines. There is also a need for more sophisticated methods for handling extreme cases of modality missingness and imbalance. The exploration of different meta-learning strategies to further enhance the adaptability and efficiency of federated multimodal learning systems remains an open and promising area of research. 

\bibliographystyle{abbrv}
\bibliography{neurips}
\newpage
\section*{Appendix}
\begin{figure}[ht]
    \centering
    \includegraphics[width=0.95\linewidth]{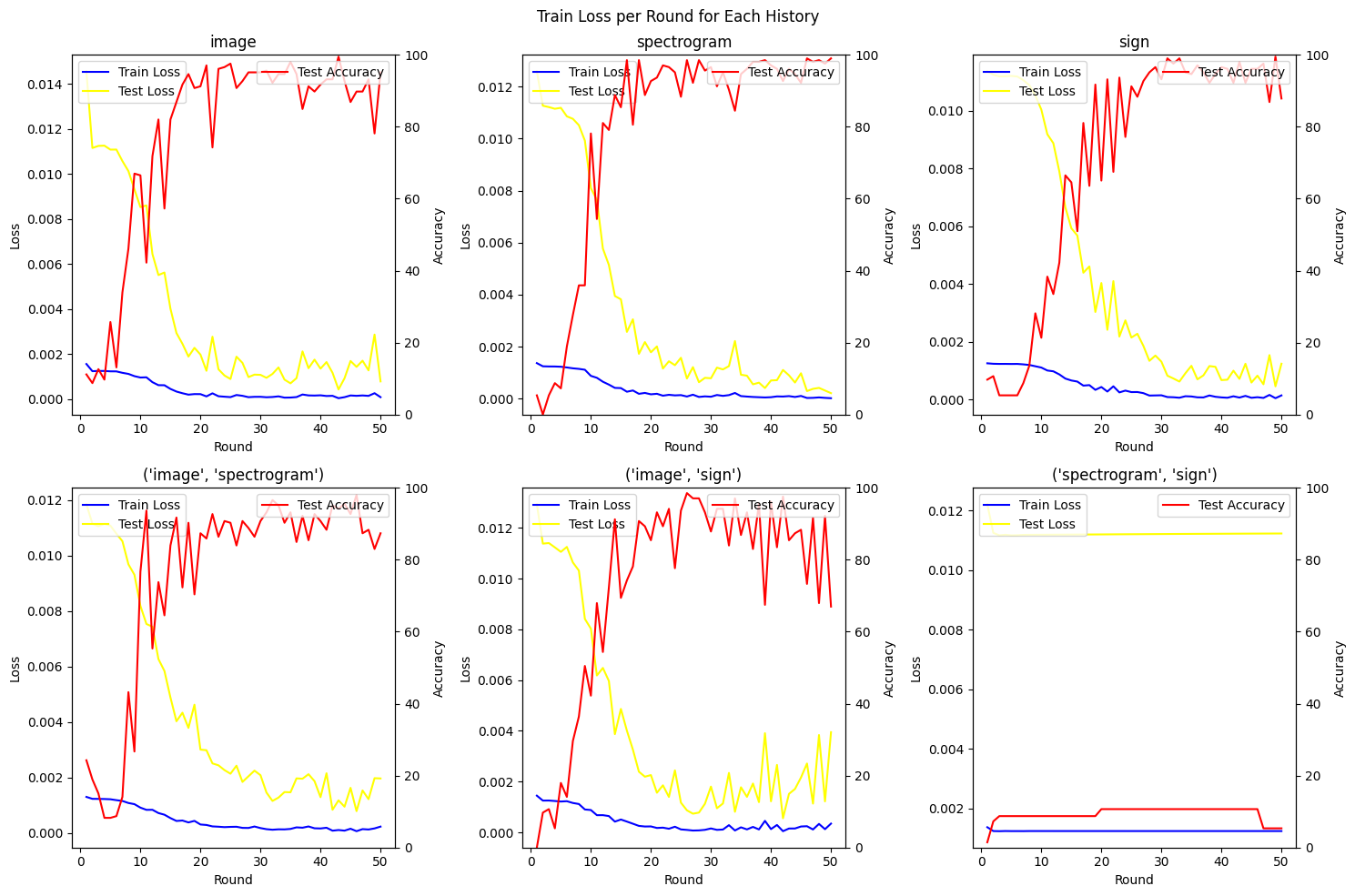}
    \caption{Train/Test with 50 communication rounds, 10 local epochs, 5 clients \#Exp 1}
    \label{5_clients_1}
\end{figure}

\begin{figure}[ht]
    \centering
    \includegraphics[width=0.95\linewidth]{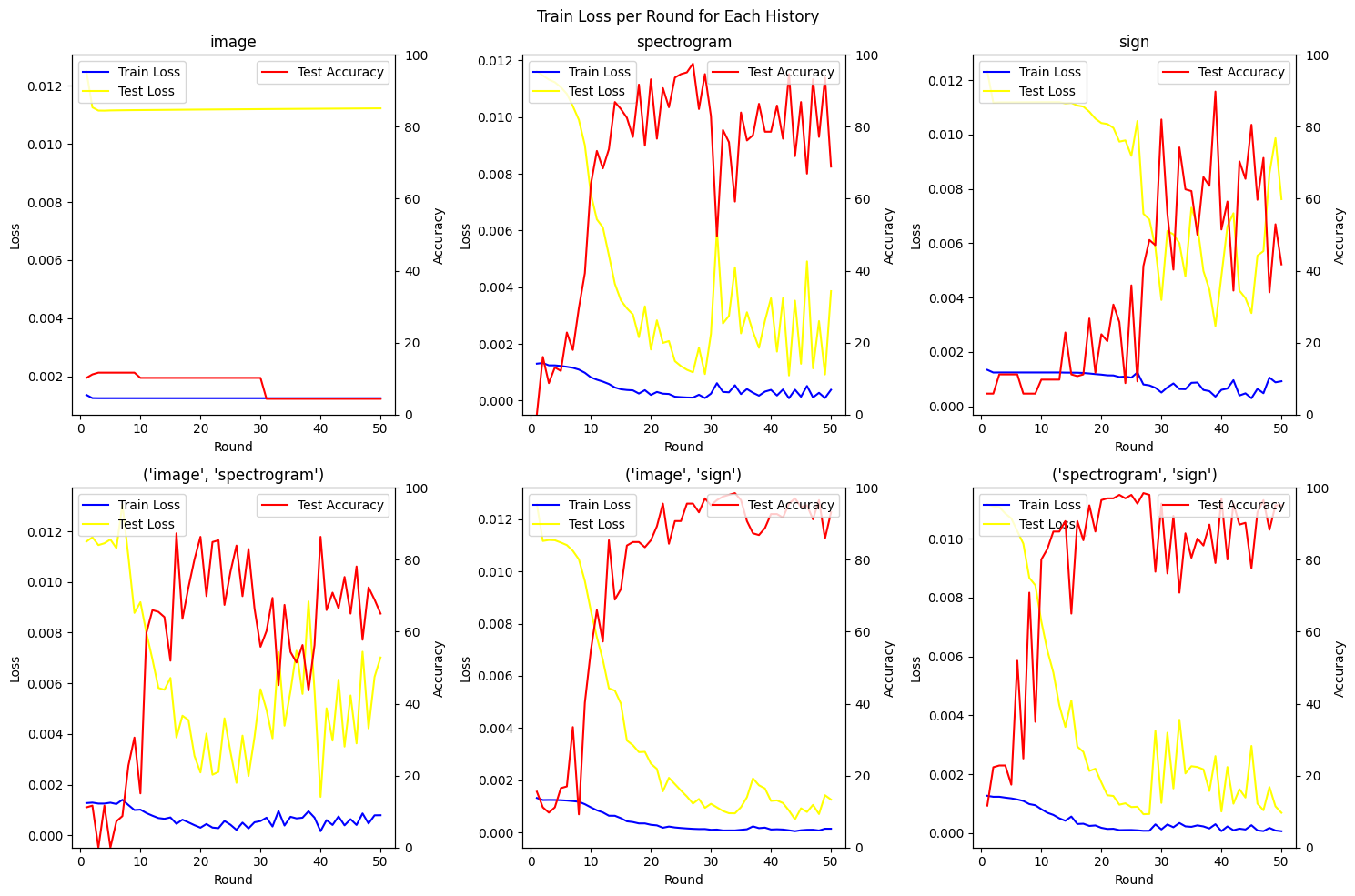}
    \caption{Train/Test with 50 communication rounds, 10 local epochs, 5 clients \#Exp 2}
    \label{5_clients_2}
\end{figure}

\begin{figure}[ht]
    \centering
    \includegraphics[width=0.95\linewidth]{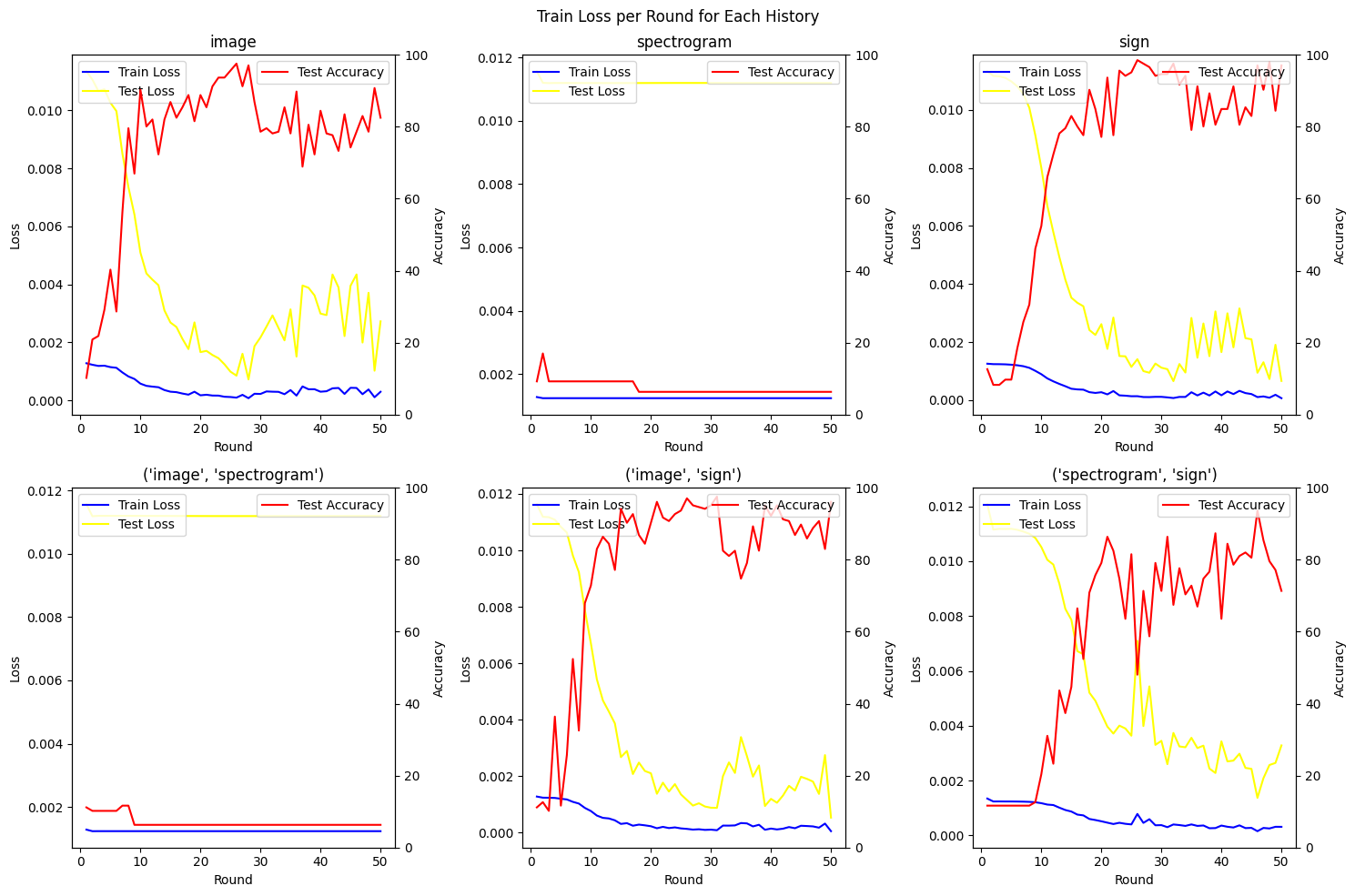}
    \caption{Train/Test with 50 communication rounds, 10 local epochs, 10 clients}
    \label{10_clients}
\end{figure}

\begin{figure}[ht]
    \centering
    \includegraphics[width=0.6\linewidth]{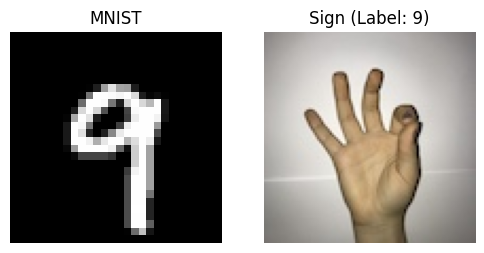}
    \caption{An example of aligned digit and sign}
    \label{aligned}
\end{figure}

\begin{figure}[ht]
    \centering
    \includegraphics[width=0.4\linewidth]{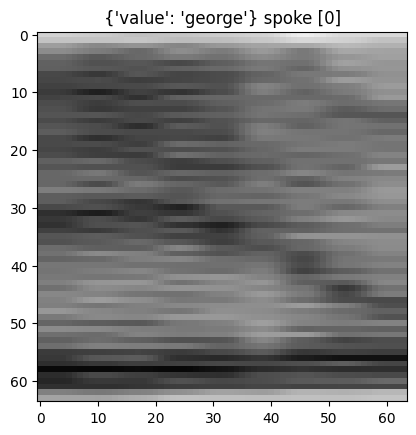}
    \caption{An example of sound represented as spectrogram}
    \label{spect}
\end{figure}

\end{document}